\documentclass{article}
\usepackage{ijcai23}

\usepackage{times}
\usepackage{soul}
\usepackage{url}
\usepackage[hidelinks]{hyperref}
\usepackage[utf8]{inputenc}
\usepackage[small]{caption}
\usepackage{graphicx}
\usepackage{amsmath}
\usepackage{amsthm}
\usepackage{booktabs}
\usepackage[switch]{lineno}

\usepackage{tikz}
\usepackage{xcolor}
\usepackage{subcaption}
\usepackage{bm}
\usepackage{amssymb}
\usepackage{mathtools}
\usepackage{multicol}
\usepackage{comment}
\DeclareMathOperator*{\argmax}{arg\,max}
\usepackage[boxed, ruled, linesnumbered]{algorithm2e}
\SetKw{KwBy}{by}
\SetKw{KwReturn}{return}

\SetCommentSty{mycommfont}
\usepackage{flushend}
\usepackage{float}

\title{BRExIt: On Opponent Modelling in Expert Iteration}

\author{
Daniel Hernandez$^{1,2}$
\and
Hendrik Baier$^3$\and
Michael Kaisers$^4$
\affiliations
$^1$Sony AI\\
$^2$University of York, UK\\
$^3$Eindhoven University of Technology, The Netherlands\\
$^4$Centrum Wiskunde \& Informatica, The Netherlands
}

\begin{document}

\maketitle

\begin{abstract}
Finding a best response policy is a central objective in game theory and multi-agent learning, with modern population-based training approaches employing reinforcement learning algorithms as best-response oracles to improve play against candidate opponents (typically previously learnt policies).
We propose Best Response Expert Iteration (BRExIt), which accelerates learning in games by incorporating opponent models into the state-of-the-art learning algorithm Expert Iteration (ExIt). BRExIt aims to (1) improve feature shaping in the apprentice, with a policy head predicting opponent policies as an auxiliary task, and (2) bias opponent moves in planning towards the given or learnt opponent model, to generate apprentice targets that better approximate a best response.
In an empirical ablation on BRExIt's algorithmic variants against a set of fixed test agents, we provide statistical evidence that BRExIt learns better performing policies than ExIt. 
\end{abstract}

\section{Introduction}\label{section:introduction}
Reinforcement learning has been successfully applied in increasingly challenging settings, with multi-agent reinforcement learning being one of the frontiers that pose open problems~\cite{hernandez2017survey,albrecht2018autonomous,nashed2022survey}. Finding strong policies for multi-agent interactions (games) requires techniques to selectively explore the space of best response policies, and techniques to learn such best response policies from gameplay. We here contribute a method that speeds up the learning of approximate best responses in games.

State-of-the-art training schemes such as population based training with centralized control~\cite{lanctot2017unified,liu2021towards} employ a combination of outer-loop training schemes and inner-loop learning agents in a nested loop fashion~\cite{hernandez2019generalized}. In the outer loop, a combination of fixed policies is chosen via game theoretical analysis to act as static opponents. Within the inner loop, a reinforcement learning (RL) agent repeatedly plays against these static opponents, in order to find an approximate best response policy against them. In theory, any arbitrary RL algorithm could be used as such a \textbf{policy improvement operator/best response oracle} in the inner loop. The best response policy found is then added to the training scheme's population, and the outer loop continues, constructing a population of increasingly stronger policies over time. This technique has been successfully applied in highly complex environments such as Dota~2~\cite{berner2019dota} and Starcraft~II ~\cite{vinyals2019grandmaster}, with PPO~\cite{Schulman2017} and IMPALA~\cite{Espeholt2018} respectively used as policy improvement operators.

The large number of training episodes required by modern deep RL (DRL) algorithms as policy improvement operators makes the inner loop of such training schemes a computational bottleneck. In this paper we accelerate approximating best responses by introducing \emph{\textbf{Best Response Expert Iteration} (BRExIt)}. We extend Expert Iteration (ExIt)~\cite{anthony2017thinking}, famously used in AlphaGo~\cite{silver2016mastering}, by introducing opponent models (OMs) in both (1) the \textbf{apprentice} (a deep neural network), with the aim of feature shaping and learning a surrogate model, and (2) the \textbf{expert} (Monte Carlo Tree Search), to bias its search towards approximate best responses to the OMs, yielding better targets for the apprentice. BRExIt thus better exploits OMs that are available in centralized training approaches, or also in Bayesian settings which assume a given set of opponents~\cite{oliehoek2014best}.

For the case of given OMs that are computationally too demanding for search, but which can be used to generate training games, we also test a variant of BRExIt that uses learned surrogate OMs instead. In the game of Connect4, we find BRExIt learns significantly stronger policies than ExIt with the same computation time, alleviating the computational bottleneck in training towards a best response.

\section{Related work}
BRExIt stands at the confluence of two streams of literature: improving the sample efficiency of the ExIt framework (Section~\ref{chapter2:section:expert_iteration_and_improvements}), and incorporating opponent models within deep reinforcement learning (Section~\ref{chapter2:section:opponent_modelling_in_drl}).

\subsection{Expert Iteration and improvements} \label{chapter2:section:expert_iteration_and_improvements}

Expert Iteration~\cite{anthony2017thinking} combines planning and learning during training, with the goal of finding a parameterized policy $\pi_{\bm{\theta}}$ (where $\bm{\theta} \in \mathbb{R}^n$) that maximizes a reward signal. The trained policy can either be used within a planning algorithm or act as a standalone module during deployment without needing environment models.

ExIt's two main components are (1) the \textbf{expert}, traditionally an MCTS procedure~\cite{browne2012survey}, and (2) the \textbf{apprentice}, a parameterized policy $\pi_{\bm{\theta}}$, usually a neural network. In short, the expert takes actions in an environment using MCTS, generating a dataset of good quality moves. The apprentice updates its parameters $\bm{\theta}$ to better predict both the expert's actions and future rewards. The expert in turn uses the apprentice inside MCTS to bias its search. Updates to the apprentice yield higher quality expert searches, yielding new and better targets for the apprentice to learn. This iterative improvement constitutes the main ExIt training loop, depicted at the top of Figure~\ref{figure:brexit_vs_exit}.

Significant effort has been aimed at improving ExIt, e.g. exploring alternate value targets~\cite{willemsen2021value} or incorporating prioritized experience replay~\cite{Schaul2015}, informed exploratory measures~\cite{soemers2020manipulating} or domain specific auxiliary tasks for the apprentice~\cite{wu2019accelerating}. BRExIt improves upon ExIt by deeply integrating opponent models within it.

\subsection{Opponent modelling in DRL}\label{chapter2:section:opponent_modelling_in_drl}

Policy reconstruction methods predict agent policies from environment observations via OMs~\cite{albrecht2018autonomous}, which has been shown to be beneficial in collaborative~\cite{carroll2019utility}, competitive~\cite{nashed2022survey} and mixed settings~\cite{hong2018deep}. Deep Reinforcement Opponent Modelling (DRON) was one of the first works combining DRL with opponent modelling~\cite{he2016opponent}. The authors used two networks, one that learns $Q$-values using Deep Q-Network (DQN)~\cite{mnih2013playing}, and another that learns an opponent's policy by observing hand-crafted opponent features. Their key innovation is to combine the output of both networks to compute a $Q$-function that is \textbf{conditioned} on (a latent encoding of) the approximated opponent's policy. This accounts for a given agent's $Q$-value dependency on the other agents' policies. Deep Policy Inference Q-Network (DPIQN)~\cite{hong2018deep} brings two further innovations: (1) merging both modules into a single neural network, and (2) baking the aforementioned $Q$ function conditioning into the neural network architecture by reusing parameters from the OM module in the $Q$ function.

OMs within MCTS have been shown to improve search~\cite{timbers2022approximate} when assuming access to ground truth opponent models, or partially correct models~\cite{goodman2020does}. As auxiliary tasks for the apprentice, OMs have been used in ExIt to predict the follow-up move an opponent would play in sequential games~\cite{wu2019accelerating}. BRExIt both learns opponent models as an auxiliary task, and uses them inside of MCTS.

\section{Background}
Section~\ref{section:marl} introduces relevant RL and game theory constructs, followed by the approach to opponent modelling used in BRExIt in Section~\ref{section:background_opponent_modelling_drl} and the inner workings of ExIt in Section~\ref{section:background_expert_iteration}.

\subsection{Multiagent Reinforcement Learning}\label{section:marl}



Let \textit{E} represent a fully observable stochastic game with $n$
agents, state space $S$, shared action space $A$ and shared policy space $\Pi$.
Policies are stochastic mappings from states to actions, with $\pi_i: S \times
A \rightarrow [0, 1]$ denoting the $i$th agent's policy, and $\boldsymbol{\pi} =
[\pi_1,\ldots,\pi_n]$ the joint policy vector, which can be regarded as a
distribution over the joint action space. $T: S \times A \times S \rightarrow
[0, 1]$ is the transition model (the environment dynamics), determining how an
environment state $s$ changes to a new state $s'$ given a joint action
$\bm{a}$. $R_i: S \times A \times S \rightarrow \mathbb{R}$ is agent $i$'s
reward function. $G^i_t \in \mathbb{R}$ is the return from time $t$ for
agent $i$, the accumulated reward obtained by agent $i$ from time $t$ until
episode termination; $\gamma \in (0, 1]$ is the environment's discount
factor. When considering the viewpoint of a specific agent $i$, we decompose a
joint action vector $\bm{a} = (a_i, \bm{a_{-i}})$ and joint policy vector
$\bm{\pi} = (\pi_i, \bm{\pi_{-i}})$ into the individual action or policy for
agent $i$, and the other agents denoted by $-i$.

The state-value function $V^{\bm{\pi}}_i: S \rightarrow \mathbb{R}$ denotes
agent $i$'s expected cumulative reward from state $s$ onwards assuming all
agents act as prescribed by $\bm{\pi}$.
\begin{align*}
    V^{\bm{\pi}}_i(s) = &\sum_{\bm{a} \in A} \bm{\pi}(\bm{a} |s) \nonumber\\
                        &\sum_{s' \in S} T(s' | s, a_i, \bm{a_{-i}})
                 [R_i(s, a_i, \bm{a_{-i}}, s') + \gamma V^{\bm{\pi}}_i(s')]
\end{align*}
Agent $i$'s optimal policy depends on the other policies:
\begin{equation}
    \pi^*_i(\cdot | s) = \argmax_{\pi_i} V^{(\pi_i, \bm{\pi_{-i}})}_i(s)
    \label{equation:multiagent_optimal_policy}
\end{equation}
Assuming $\bm{\pi_{-i}}$ to be stationary, agent $i$'s optimal policy $\pi^*_i$
is also called a \textbf{best response} $\pi^*_i \in BR(\bm{\pi_{-i}})$, where
$BR(\bm{\pi_{-i}})$ denotes the set of all best responses against
$\bm{\pi_{-i}}$. Our goal is to train a system to (1) predict and encode what is knowable about the opponents' policies $\bm{\pi_{-i}}$ and (2) compute a best response to that. 

\subsection{Opponent modelling in DRL}\label{section:background_opponent_modelling_drl}

We follow the opponent modelling approach popularized by
DPIQN~\cite{hong2018deep}, which uses a neural network to both learn opponent
models and an optimal $Q$-function. The latter is learnt by minimising the loss function $\mathcal{L}_{Q}$ of DQN~\cite{mnih2013playing}. Opponent modelling is an auxiliary task, trained by minimising the cross-entropy between
one-hot encoded observed action for each agent $j$, $\bm{a}^j$, and their
corresponding predicted opponent policies at state $s$, $\hat{\pi}_j(\cdot |
s)$, defined as the \textbf{policy inference} loss $\mathcal{L}_{PI}$ in
Equation~\ref{equation:dpiqn_total_loss}. These two losses are combined into $\mathcal{L}_{DPIQN}$ with an adaptive weight to improve learning stability.
\begin{subequations}
\begin{align}
  \mathcal{L}_{PI} &= - \frac{1}{N} \sum_{j = 0}^{N} \bm{a_j} log(\hat{\pi}_j(\cdot | s))\\
  \mathcal{L}_{DPIQN} &= \frac{1}{\sqrt{\mathcal{L}_{PI}}} \mathcal{L}_{Q} + \mathcal{L}_{PI}
\end{align}%
\label{equation:dpiqn_total_loss}%
\end{subequations}%
BRExIt makes use of both the policy inference loss for its OMs and the adaptive learning weight to regularize its critic loss. However, instead of learning a $Q$-function as a critic, BRExIt learns a state-value function $V$, as suggested by previous work~\cite{hernandez2019agent}.

\subsection{Expert Iteration}\label{section:background_expert_iteration}

We use an open-loop MCTS implementation~\cite{Silver2017b} as the expert; tree
nodes represent environment states $s$, and edges $(s, a)$ represent taking
action $a$ at state $s$. Each edge stores a set of statistics:
\begin{equation}
    \{N(s, a), Q(s, a), P(s, a), i, A_n\}
    \label{equation:mcts_node_statistics}
\end{equation}
$N(s, a)$ is the number of visits to edge $(s,a)$. $Q(s, a)$ is the mean action-value
for $(s, a)$, aggregated from all simulations that have traversed $(s, a)$.
$P(s, a)$ is the prior probability of following edge $(s, a)$. ExIt uses the
apprentice policy to compute priors, $P(s, a) = \pi_{\bm{\theta}}(a | s)$. One
of our contributions is adding opponent-awareness to the computation of these priors.
Finally, index $i$ denotes the player to act in the node and $A_n$ indicates the
available actions.

For every state $s$ encountered by an ExIt agent during a training episode, the expert takes an action $a$
computed by running MCTS from state $s$ and selecting the action
of to the root node's most visited edge $(s, a)$. During search, the
tree is traversed using the same selection strategy as
AlphaZero~\cite{Silver2017b}. Edges $(s, a)$ are traversed following the most
promising action according to the PUCT formula:
\begin{equation}
    \argmax_{a} \; Q(s, a) + C_{PUCT} \frac{P(s, a) \sqrt{\sum_{a'} N(s, a')}}{1 + N(s, a)},
    \label{equation:puct}
\end{equation}
where $C_{PUCT}$ is a tunable constant. The apprentice $\pi_{\bm{\theta}}$ is a distillation of previous MCTS searches,
and provides $P(s, a)$, thus biasing MCTS towards actions that were
previously computed to be promising. Upon reaching a leaf node with state $s'$,
we backpropagate a value given by a learnt state value function
$V_{\bm{\phi}}(s')$ with parameters $\bm{\phi} \in \mathbb{R}^n$ trained to regress against observed returns.

After completing search from a root node representing
state $s$, the policy $\pi_{MCTS}(\cdot|s)$ can be extracted from the statistics stored on
the root node's edges. This policy is stored as a training target for the
apprentice's policy head to imitate:
\begin{equation}
    \pi_{MCTS}(a | s) = \frac{N(s,a)}{\sum_{a'} N(s, a')}
    \label{equation:pi_mcts}
\end{equation}
As shown in the top right corner of Figure~\ref{figure:brexit_vs_exit} ExIt
builds a dataset containing a datapoint for each timestep $t$:
\begin{equation}
    \{s_t, \pi_{MCTS}(\cdot | s_t), G^i_t\}
    \label{equation:exit_dataset}
\end{equation}
The top left corner of Figure~\ref{figure:brexit_vs_exit} shows an actor-critic
architecture, used as the apprentice, with a policy head $\pi_{\bm{\theta}}$, and a
value head $V_{\bm{\phi}}$. A cross-entropy loss is used to train the actor
towards imitating the expert's moves, and a mean-square error loss is used to update the
critic's state value function towards observed returns $G^i_t$.

\section{BRExIt: Opponent modelling in ExIt}\label{section:opponent_modelling_in_exit}
\begin{figure*}[htb]
    \centering
    \includegraphics[width=0.9\textwidth]{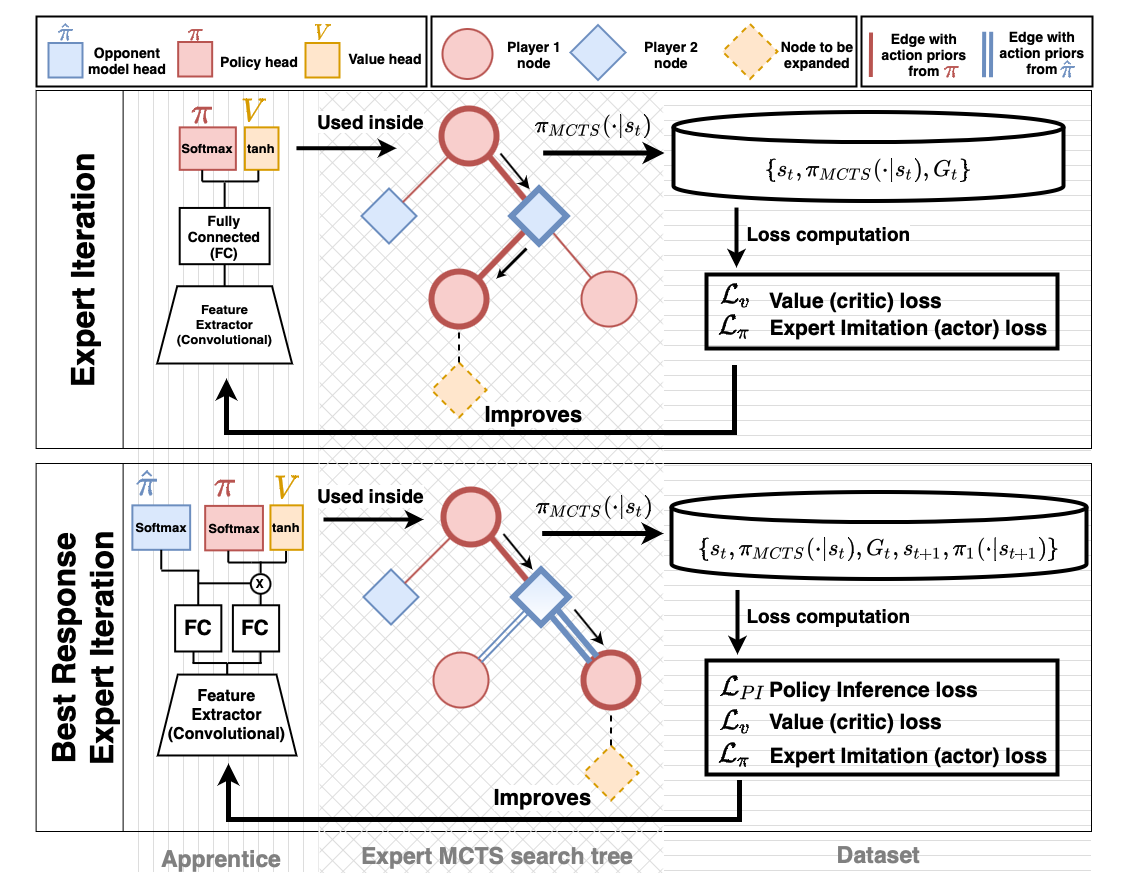}
\caption{Illustration of ExIt (top half) and BRExIt (bottom half) for a
2-player game. BRExIt bases the decision on how to compute edge priors based on
whose turn a node corresponds to. For opponent nodes, during training, these
action priors can come from the true opponent's policy or from the apprentice's
opponent model heads $\textcolor{blue}{\hat{\pi}}$. BRExIt adds an opponent
modelling loss by also gathering the states observed by the opponent $s_{t+1}$
and either the output of their policy $\pi_1(\cdot|s_{t+1})$, or
a corresponding one-hot encoded action.}
    \label{figure:brexit_vs_exit}
\end{figure*}

We present Best Response Expert Iteration (BRExIt), an extension on ExIt that uses opponent modelling for two purposes. First, to enhance the apprentice's architecture with opponent modelling heads, acting as feature shaping mechanisms -- see Subsection~\ref{subsection:opponent_models_in_apprentice}. Second, to allow the MCTS expert to approximate a best response against a set of opponents: $\pi_{MCTS} \in BR(\bm{\pi_{-i}})$ -- see Subsection~\ref{subsection:opponent_modelling_inside_mcts}. We visually compare BRExIt and ExIt in Figure~\ref{figure:brexit_vs_exit}.

\subsection{Learning opponent models in sequential games}\label{subsection:opponent_models_in_apprentice}

Previous approaches to learn opponent models used observed actions as learning
targets, coming from a game theoretical tradition where individual actions can
be observed but not the policy that generated them~\cite{brown1951iterative}.
However, the centralized population-based training schemes which motivate our
research already require access during training to all policies in the
environment, both for the training agent and the opponents' policies. We
further exploit this assumption by separately testing two options for the learning targets in the policy inference loss from Equation~\ref{equation:dpiqn_total_loss}: the one-hot encoded observed opponent actions, or the full distributions over actions computed by the opponents during play.

Prior work has typically focused on fully observable
\textit{simultaneous} games, where a shared environment state is used by all
agents to compute an action at every timestep. Thus, if agent $i$ wanted to
learn models of its opponents' policies, storing (a) each of $i$'s observed
shared states and (b) opponent actions, was sufficient to learn opponent
models. We extend this to sequential games, where agents take turns acting
based on individual states, in the following way. BRExIt augments the dataset
collected by ExIt, specified in Equation~\ref{equation:exit_dataset}, by adding
(1) the state which each opponent agent $j \neq i$ observed and (2) either only the observed opponent actions or the full ground truth action distributions given by the opponent policies in their
corresponding turn. The data collection process for sequential environments is
described in Algorithm~\ref{algorithm:brexit_collection_sequential}. Formally, for the BRExIt agent acting at timestep $t$ and its $n_o$ opponents acting at timesteps $t+1$ to $t+n_o$, BRExIt adds to its dataset either the observed action of every agent $j$, or their policy $\pi_j$ evaluated at the state of their turn.
\begin{equation}
    \{s_t, \pi_{MCTS}(\cdot | s_t), \{s_{t+j}, \pi_j(\cdot | s_{t+j})\}^{n_o}_{j=1}, G^i_t\}
    \label{equation:brexit_dataset}
\end{equation}
\subsection{Apprentice Representation}\label{section:apprentice_representation}

BRExIt's three headed apprentice architecture extends ExIt's
actor-critic representation with opponent models as per DPIQN's
design~\cite{hernandez2019agent}, depicted on the bottom left of
Figure~\ref{figure:brexit_vs_exit} for a single opponent. This architecture
reuses parameters from the OM as a feature shaping mechanism for both the
actor and the critic. It takes as input an environment state $s_t
\in \mathcal{S}$ for a timestep $t$, which can correspond to the observed state
of any agent. It features 3 outputs: (1) the apprentice's actor policy
$\pi_{\bm{\theta}}(s_t)$ (2) the state-value critic $V^{\pi_{\bm{\theta}},
\bm{\hat{\pi}_{\Psi}}}_{\bm{\phi}}(s_t)$ (3) the opponent models
$\hat{\pi}^{\bm{\psi}_j}(s_t) \in \bm{\hat{\pi}^{\Psi}}$, where $\bm{\Psi}$
contains the parameters for all opponent models and $\psi_j \subset \bm{\Psi}$ the
parameters for opponent model head $j$.

On certain sequential games the distribution of states encountered by each
agent might differ, and so the actor-critic head and each of the OM
heads could each be trained on different state distributions. In practice this
means that the output of one of these heads might only be usable if the input
state $s_t$ for a timestep $t$ comes from the distribution it was trained on; however, the OM can in any case help via feature shaping. This does not apply to simultaneous games where all agents observe the same state.

\subsection{Opponent modelling inside MCTS}~\label{subsection:opponent_modelling_inside_mcts}
\noindent BRExIt follows
Equation~\ref{equation:brexit_action_priors_computation} to compute
the action priors $P(s, a)$ from Equation~\ref{equation:puct} for a node with player index $j$, notably
using opponent models in nodes within the search tree that correspond to
opponents' turns. This process is exemplified in the lower middle half of
Figure~\ref{figure:brexit_vs_exit}. Such opponent models can be either
the ground truth opponent policies (the real opponent policies $\bm{\pi_{-i}}$)
by exploiting centralized training scheme assumptions, or otherwise the
apprentice's learnt opponent models $\bm{\hat{\pi}^{\psi}}$. This is a key
difference from ExIt, which always uses the apprentice's
$\pi_{\bm{\theta}}$ to compute $P(s, a)$.
\begin{equation}
    \label{equation:brexit_action_priors_computation}
    P(s, a) = 
    \begin{cases*}
        \pi_{\bm{\theta_{\phantom{j}}}}(a | s) & $j$ == BRExIt player index \\
        \pi^j_{-i}(a | s) & For ground truth models \\
        \hat{\pi}^{\bm{\psi_j}}(a | s) & For learnt opponent models
    \end{cases*}
\end{equation}
By using either the ground truth or learnt opponent models, we initially bias
the search towards a best response against the actual policies in the
environment. However, note that initial
$P(s, a)$ values will be overridden by the aggregated simulation returns $Q(s,
a)$, as with infinite compute MCTS converges to best response against a
perfectly rational player, whose actions may deviate from the underlying
opponent's policy. Thus, BRExIt's search maintains the asymptotic behaviour of MCTS
while simultaneously priming the construction of the tree towards areas which
are likely to be explored by the policies in the environment.

In contrast to BRExIt, ExIt biases its expert search towards a best response against the apprentice's own policy, by using the apprentice $\pi_{\bm{\theta}}$ to compute $P(s, a)$ at \textit{every} node in the tree. MCTS
here assumes that all agents follow the same policy as the apprentice, whereas in reality agents might follow any arbitrary policy. Not trying to exploit the opponents it is trained against, ExIt generates more conservative searches. Recent studies show that this conservativeness can be detrimental in terms of finding diverse sets of policies throughout training for population-based training schemes~\cite{balduzzi2019open,liu2021towards}. Instead, they advocate for a more direct computation of best responses against known opponents as a means to discover a wider area of the policy space. This allows the higher level training scheme to better decide on which opponents to use as targets to guide future exploration. We argue that BRExIt has this property built-in, by actively biasing its search towards a best response against ground truth opponent policies. Future work could investigate this claim.

We could have designed BRExIt to be even more exploitative towards opponent models, by masking actions sampled from these models as part of the environment dynamics. While this would yield actions very specifically targeted to respond to the modelled policies, it makes such best reponses very brittle, and can be problematic especially for imperfect OMs. In contrast, we propose using opponent models as priors in BRExIt, such that planning can still improve upon the opponent policies; this results in more robust learning targets.

Algorithms~\ref{algorithm:brexit_collection_sequential},
\ref{algorithm:brexit_update_sequential} and \ref{algorithm:brexit_training}
depict BRExIt's data collection, model update logic and overarching training
loop respectively for a sequential environment. Coloured lines represent our
contributions w.r.t ExIt.

\begin{algorithm}
    \KwIn{$(\text{apprentice } \pi_{\bm{\theta}}, \textcolor{orange}{\text{opp. models } \hat{\bm{\pi}}^{\bm{\psi}}_{-i}}, \text{critic } V_{\bm{\phi}})$}
    \KwIn{\textcolor{orange}{\textit{Opponent policies}: $\bm{\pi_{-i}}$}}
    \KwIn{Environment: $E = (\mathcal{P}, \rho_0)$}
    Initialize dataset: $D = [\ ]$\;
    Initialize time $t \leftarrow 0$\;
    Sample initial state $s_0 \sim \rho_0$\;
    \While{$s_t$ \text{is not} $terminal$} {
       Search: $a_t, tree = MCTS(s_t, \pi_{\bm{\theta}}, \textcolor{orange}{\bm{\pi_{-i}}}, V_{\bm{\psi}})$\;
       Act in the game $s_{t+1}, r_t \sim \mathcal{P}(s_t, a_t)$\; 
       Get from $tree$: $\pi_{MCTS}(s_t, a) = \frac{N_{root}(s_t,a)}{\sum_{a'} N_{root}(s_t, a')}$\;
       \For{$j = 1,\ldots,|\bm{\pi_{-i}|}$} {
           Sample opp. action: $a_{t+j} \sim \bm{\pi_{-i}}^j(s_{t+j})$\;
           Act in the game $s_{t+j}, \_ \sim \mathcal{P}(s_{t+j}, a_{t + j})$
       }
       $D \cup \{s_t, \pi_{mcts}(s_t),$ \textcolor{orange}{$\{s_{t+j}, \bm{\pi_{-i}}^j(s_{t+j})\}^{|\bm{\pi_o}|}_{j=1}$}$, r_i\}$\;
       $t \leftarrow t + |\bm{\pi_{-i}}|$\;
    }
    \KwReturn $D$\;
    \caption{BRExIt data collection}
    \label{algorithm:brexit_collection_sequential}
\end{algorithm}

\begin{algorithm}
\KwIn{\textit{Three head network}: $NN = (\pi_{\bm{\theta}}, \textcolor{orange}{\hat{\bm{\pi}}^{\bm{\psi}}_{-i}}, V_{\bm{\phi}})$}
\KwIn{\textit{Dataset}: $D$}
\For{$t = 0, 1, 2, \dots$} {

   Sample $n$ datapoints from $D$: \\ $(s_t, r_t, \pi_{MCTS}(s_t, \cdot), r_i, \linebreak \textcolor{orange}{\{s_{t+j}, \bm{\pi_{-i}}^j(\cdot | s_{t+1})\}^{|\hat{\bm{\pi}}^{\bm{\psi}}_{-i}|}_{j=1}})_{1, \ldots, n}$\;

   MSE value loss: $\mathcal{L}_{v} = (v - V_{\bm{\phi}}(s_t))^2$ \;

   CE policy loss $\mathcal{L}_{\pi} = \pi_{MCTS}(s_t) \log{(\pi_{\bm{\theta}}(s_t))}$\;

   \textcolor{orange}{CE policy inference loss: $\mathcal{L}_{PI} = \frac{1}{|\bm{\pi}_{-i}|} \sum^{|\bm{\pi_{-i}}|}_{j=1} \pi^{j}_{-i}(s_{t+j}) \log{(\hat{\pi}^{\bm{\psi_j}}_{-i}(s_{t+j}))}$}\;

   \textcolor{orange}{Policy inference weight: $\lambda = \frac{1}{\sqrt{\mathcal{L}_{PI}}}$}\;

   Weighted final loss $\mathcal{L}_{total} = \textcolor{orange}{\lambda} (\mathcal{L}_{v} + \mathcal{L}_{\pi}) + \textcolor{orange}{\mathcal{L}_{PI}}$\;

   Backpropagate  $\nabla\mathcal{L}_{total}$ through $\bm{\theta}, \textcolor{orange}{\bm{\psi}}, \bm{\phi}$ \;
}

\caption{BRExIt model update}\label{algorithm:brexit_update_sequential}
\end{algorithm}

\begin{algorithm}
    \KwIn{\textit{Three head network}: $NN = (\pi_{\bm{\theta}}, \textcolor{orange}{\hat{\bm{\pi}}^{\bm{\psi}}_{-i}}, V_{\bm{\phi}})$}
    \KwIn{\textcolor{orange}{\textit{Opponent policies}: $\bm{\pi}_{-i}$}}

    \For{$training\ iteration = 0, 1, 2, \dots$} {
        Algo.~\ref{algorithm:brexit_collection_sequential}: $D = Dataset Collection(NN, \textcolor{orange}{\bm{\pi}_{-i}})$\;
        Algo.~\ref{algorithm:brexit_update_sequential}: $NN = UpdateApprentice(NN, D)$\;
    }
    \KwReturn NN
    \caption{BRExIt training loop}
    \label{algorithm:brexit_training}
\end{algorithm}

\section{Experiments \& Discussion}\label{section:experiments}
We are trying to answer the following two questions: Primarily, is BRExIt more performant than ExIt at distilling a competitive policy against fixed opponents into its apprentice? Secondarily, are full distribution targets for learning opponent models preferable over one-hot action encodings?

\textbf{The environment:} We conducted our experiments in the fully observable, sequential two-player game of Connect4, which is computational amenable and possesses a high degree of skill transitivity~\cite{czarnecki2020real}. We decided on using a single environment in order to obtain statistically significant results through a larger number of runs over granular algorithmic ablations. We acknowledge the limitations of using a single test domain.

\textbf{Test opponents:} We generated two test agents $\pi_{weak}, \pi_{strong}$ by freezing copies of a PPO~\cite{Schulman2017} agent trained under $\delta=0$-Uniform self-play~\cite{hernandez2019generalized} after 200k and 600k episodes. Motivated by population-based training schemes, we also used an additional opponent policy $\pi_{mixed}$, which randomly selects one of the test agents every episode.

\textbf{Trained agents:} We independently trained 7 types of agents for 48 wall-clock hours each, performing an additive construction from ExIt to BRExIt. \textbf{ExIt} is the original algorithm, \textbf{ExIt-OMFS} denotes ExIt using OMs \textit{only} for feature shaping. \textbf{BRExIt-OMS} additionally uses learnt OMs during search and \textbf{BRExIt} uses the ground truth OMs during search. For the agents using OMs, we trained both a version using full action distributions as action targets and another with one-hot encoded action targets. Each algorithm was independently trained 10 times against the 3 test opponents, yielding a total of 280 training runs. Following statistical practices~\cite{agarwal2021deep} we use Inter Quartile Metrics (IQM) for all results, discarding the worst and best performing 25\% runs to obtain performance metrics less susceptible to outliers.

\subsection{On BRExIt's performance vs. ExIt}

\begin{figure}[H]
  \begin{subfigure}[t]{\columnwidth}
    \includegraphics[width=\textwidth, keepaspectratio]{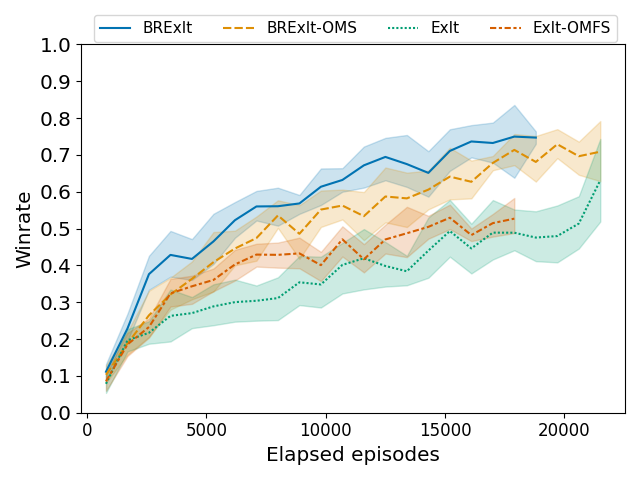}
    \caption{VS Weak agent}
    \label{figure:apprentice_convergence_ablation_weak}
  \end{subfigure}
  \begin{subfigure}[t]{\columnwidth}
    \includegraphics[width=\columnwidth, keepaspectratio]{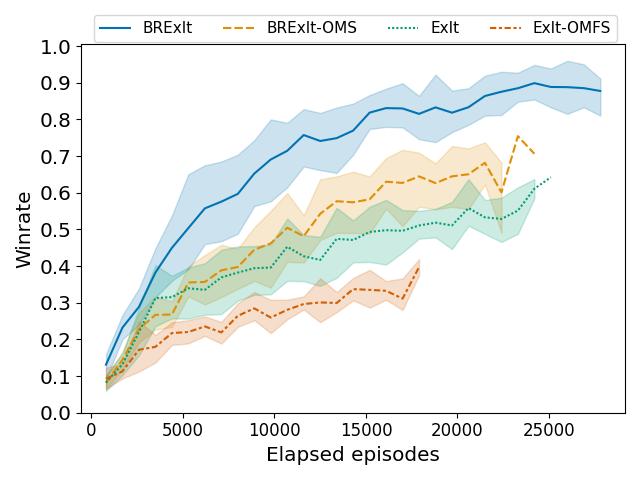}
    \caption{VS Strong}
    \label{figure:apprentice_convergence_ablation_strong}
  \end{subfigure}
  \begin{subfigure}[t]{\columnwidth}
    \includegraphics[width=\columnwidth, keepaspectratio]{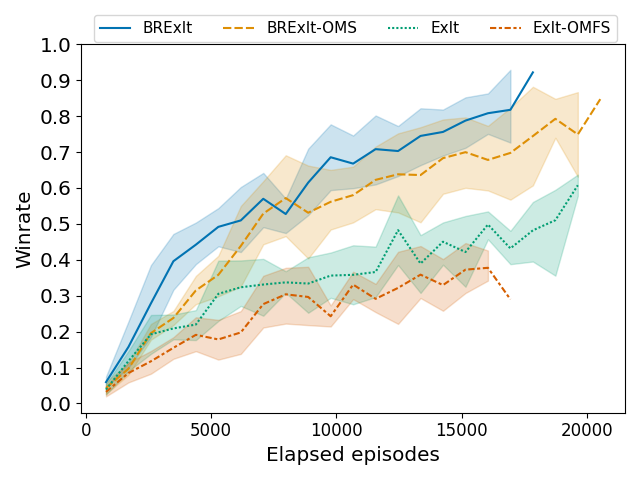}
    \caption{VS Mixed}
    \label{figure:apprentice_convergence_ablation_mixed}
  \end{subfigure}
\caption{The evolution of winrates for each ablation during training vs fixed opponents for 48h wall-clock time. Lines represent the mean value the of final apprentice's winrate over all runs. Higher is better; shaded areas show 95\% bootstrap confidence intervals.}
\label{fig:winrate_evolution}
\end{figure}

To answer our first question, Figure~\ref{fig:winrate_evolution} shows the evolution of the winrate of each of the ablation's apprentice policies throughout training. A datapoint was computed every policy update (i.e every 800 episodes) by evaluating the winrate of the apprentice policy against the opponent over 100 episodes. (Note that the difference in number of episodes between all

\begin{figure}[htb]
\centering
    \includegraphics[width=0.85\columnwidth]{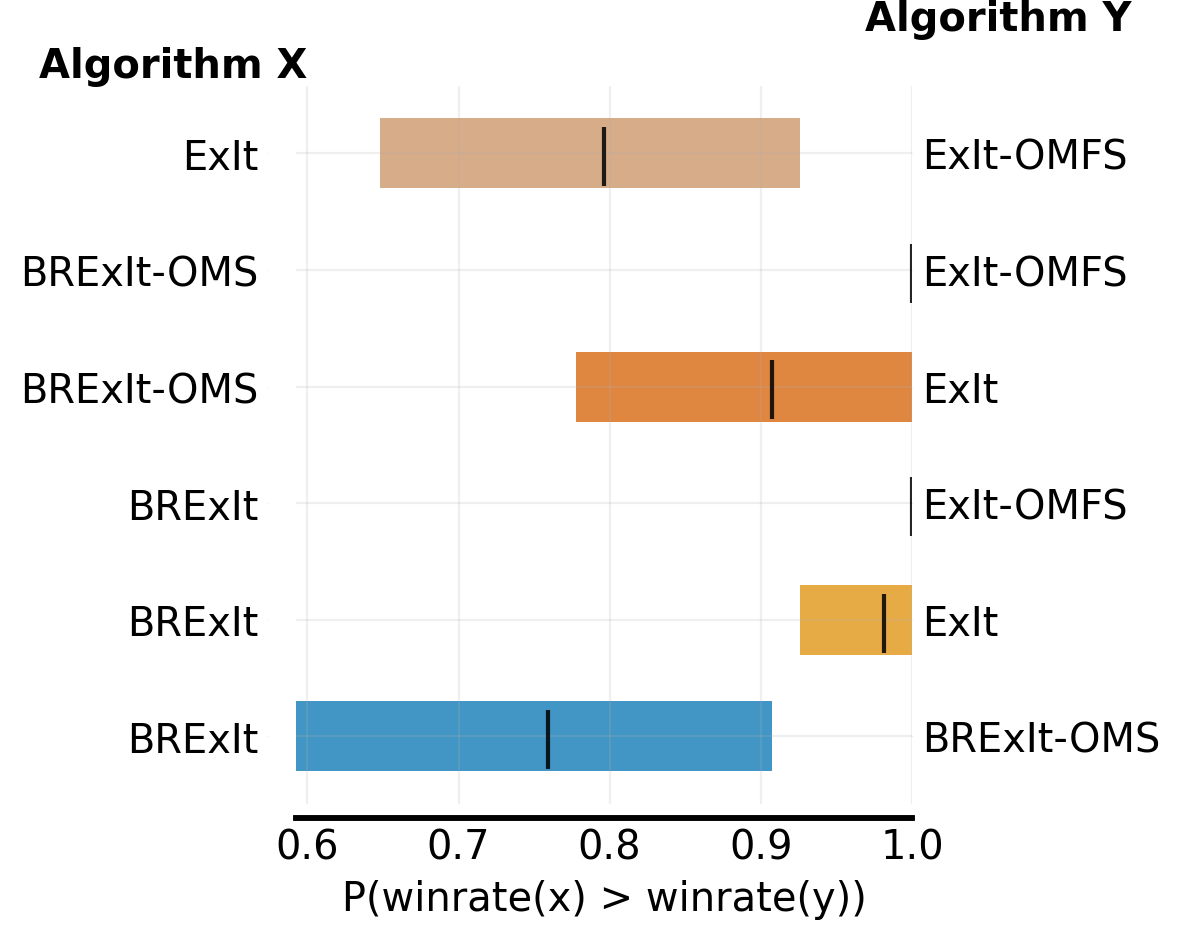}
    \caption{Each row shows the probability (vertical marker) that the algorithm $X$ (left) trains its apprentice's policy to reach a higher winrate than algorithm $Y$ (right) after the allotted 48h. Colored bars indicate 95\% bootstrap confidence intervals. Note that every training run for both BRExIt and BRExIt-OMS yielded higher performing policies than any run from ExIt-OMFS, and thus their comparisons have a 100\% probability of improvement over ExIt-OMFS.}
    \label{figure:probability_of_improvement}
\end{figure}

  \noindent ablations depends on the average episode length over the 48h of training time, which can vary as a function of both players involved.) Figure~\ref{figure:probability_of_improvement} analyzes these results and shows the probability of improvement (PoI) that one ablation has over another, defined as the probability that algorithm $X$ would yield an apprentice policy which has a higher winrate against its training opponent that algorithm $Y$~\cite{agarwal2021deep}. All agents are using full distribution OM targets here. 


Figure~\ref{fig:winrate_evolution} shows that BRExIt style agents consistently achieve a higher winrate than ExIt agents. BRExIt regularly outperforms BRExIt-OMS (77\% PoI), successfully exploiting the centralized assumption of having opponent policies available during search for extra performance. If opponent policies can only be sampled for training games but not during search, using learnt OMs during search is still beneficial, as we see that BRExIt-OMS consistently outperforms ExIt (90 \% PoI). Surprisingly, ExIt-OMFS performs worse than ExIt by a significant margin (the latter has a 80\% PoI against the former), providing empirical evidence that OMs with static opponents can be \textit{detrimental} for ExIt if OMs are not exploited within MCTS. This goes against previous results~\cite{wu2019accelerating}, which explored OMs within ExIt merely as a feature shaping mechanism and claimed modest improvements when predicting the opponent's follow-up move. Differences may be attributed to the fact that we model the opponent policy on the current state, instead of the next state.

In summary, with BRExIt (using ground truth OMs) and BRExIt-OMS (using learnt OMs) featuring a $>97\%$ and $>91\%$ PoI respectively against vanilla ExIt, our empirical results warrant the use of our novel algorithmic variants instead of ExIt whenever opponent policies are available for training.

\subsection{On full distribution VS one-hot targets}

To answer our second question, we conducted Kolmogorov-Smirnov tests comparing full action distribution targets to one-hot encoded action targets for OMs. The samples we compared were the sets of winrates at the end of training, one datapoint per training run. Table~\ref{table:p_values_kolmogorov_smirnov_tests} shows the results: There is no statistical difference between agents trained with algorithms using one-hot encoded OM targets when compared to using full action distributions.  We obtain $p \gg 0.05$ for each algorithmic combination, so we cannot reject the null hypotheses that both data samples come from the same distribution; the only exception being ExIt-OMFS, which shows a statistically significant decrease in performance when using one-hot encoded targets.

These results run contrary to the intuition that richer targets for opponent models will in turn improve the quality of the apprentice's policy. However, we observed that full distributional targets do yield OMs with better prediction capabilities, as indicated by lower loss of the OM during training. This hints at the possibility that OM's usefulness increases only up to a certain degree of accuracy, echoing recent findings~\cite{goodman2020does}. Hence, while BRExIt does require access to ground truth policies during search, search in BRExIt-OMS achieves similar performance with opponent models trained on action observations, which is promising for transferring BRExIt-OMS into practical applications.

\begin{table}[htb]
    \centering
    \caption{Testing potential improvements of full distribution targets to one-hot encoded action targets. $p$-values are from two-sample Kolmogorov-Smirnov tests.}
    \begin{tabular}{l c c c}
        \toprule
        Base Algorithm & Opponent & $p$-value & Distr. targets\\
        & & & signif. better?\\
        \midrule
        BRExIt & Weak & 0.930 & No\\
        BRExIt-OMS & Weak & 0.931 & No\\
        ExIt-OMFS & Weak & 0.930 & No\\
        BRExIt & Strong & 0.930 & No\\
        BRExIt-OMS & Strong & 0.999 & No\\
        ExIt-OMFS & Strong & 0.930 & No\\
        BRExIt & Mixed & 0.142 & No\\
        BRExIt-OMS & Mixed & 0.930 & No\\
        ExIt-OMFS & Mixed & 0.025 & Yes \\
        \bottomrule
    \end{tabular}
    \label{table:p_values_kolmogorov_smirnov_tests}
\end{table}

\section{Conclusion}\label{section:Discussion}

We investigated the use of opponent modelling within the ExIt framework, introducing the BRExIt algorithm. BRExIt augments ExIt by introducing opponent models both within the \textbf{expert planning phase}, biasing its search towards a best response against the opponent, and within the \textbf{apprentice's model}, to use opponent modelling as an auxiliary task. In Connect4, we demonstrate BRExIt's improved performance compared to ExIt when training policies against fixed agents.

There are multiple avenues for future work. At the level of population based training schemes (such as self-play), future work can focus on measuring whether the quality of populations generated by different training schemes using BRExIt surpasses that of populations where ExIt is used as a policy improvement operator. In addition, search methods can struggle with complex games due to their high branching factor -- simultaneous move games for example have combinatorial action spaces -- which could be alleviated by using BRExIt's opponent models to narrow down the search space.

\bibliographystyle{named}
\bibliography{biblio}

\begin{thebibliography}{}

\bibitem[\protect\citeauthoryear{Agarwal \bgroup \em et al.\egroup
  }{2021}]{agarwal2021deep}
Rishabh Agarwal, Max Schwarzer, Pablo~Samuel Castro, et~al.
\newblock Deep reinforcement learning at the edge of the statistical precipice.
\newblock {\em Advances in Neural Information Processing Systems}, 34, 2021.

\bibitem[\protect\citeauthoryear{Albrecht and
  Stone}{2018}]{albrecht2018autonomous}
Stefano~V Albrecht and Peter Stone.
\newblock Autonomous agents modelling other agents: A comprehensive survey and
  open problems.
\newblock {\em Artificial Intelligence}, 258:66--95, 2018.

\bibitem[\protect\citeauthoryear{Anthony \bgroup \em et al.\egroup
  }{2017}]{anthony2017thinking}
Thomas Anthony, Zheng Tian, and David Barber.
\newblock Thinking fast and slow with deep learning and tree search.
\newblock In {\em NIPS}, 2017.

\bibitem[\protect\citeauthoryear{Balduzzi \bgroup \em et al.\egroup
  }{2019}]{balduzzi2019open}
David Balduzzi, Marta Garnelo, Yoram Bachrach, et~al.
\newblock Open-ended learning in symmetric zero-sum games.
\newblock In {\em International Conference on Machine Learning}, pages
  434--443. PMLR, 2019.

\bibitem[\protect\citeauthoryear{Berner \bgroup \em et al.\egroup
  }{2019}]{berner2019dota}
Christopher Berner, Greg Brockman, Brooke Chan, et~al.
\newblock Dota 2 with large scale deep reinforcement learning.
\newblock {\em arXiv preprint arXiv:1912.06680}, 2019.

\bibitem[\protect\citeauthoryear{Brown}{1951}]{brown1951iterative}
George~W Brown.
\newblock Iterative solution of games by fictitious play.
\newblock {\em Activity analysis of production and allocation}, 13(1):374--376,
  1951.

\bibitem[\protect\citeauthoryear{Browne \bgroup \em et al.\egroup
  }{2012}]{browne2012survey}
Cameron~B Browne, Edward Powley, Daniel Whitehouse, et~al.
\newblock {A survey of Monte Carlo Tree Search methods}.
\newblock {\em IEEE Transactions on Computational Intelligence and AI in
  Games}, 4(1):1--43, 2012.

\bibitem[\protect\citeauthoryear{Carroll \bgroup \em et al.\egroup
  }{2019}]{carroll2019utility}
Micah Carroll, Rohin Shah, Mark~K Ho, et~al.
\newblock On the utility of learning about humans for human-ai coordination.
\newblock In {\em Advances in Neural Information Processing Systems}, pages
  5175--5186, 2019.

\bibitem[\protect\citeauthoryear{Czarnecki \bgroup \em et al.\egroup
  }{2020}]{czarnecki2020real}
Wojciech~M Czarnecki, Gauthier Gidel, Brendan Tracey, et~al.
\newblock Real world games look like spinning tops.
\newblock {\em Advances in Neural Information Processing Systems},
  33:17443--17454, 2020.

\bibitem[\protect\citeauthoryear{Espeholt \bgroup \em et al.\egroup
  }{2018}]{Espeholt2018}
Lasse Espeholt, Hubert Soyer, Remi Munos, et~al.
\newblock {IMPALA: Scalable Distributed Deep-RL with Importance Weighted
  Actor-Learner Architectures}.
\newblock {\em CoRR}, abs/1802.01561, 2018.

\bibitem[\protect\citeauthoryear{Goodman and Lucas}{2020}]{goodman2020does}
James Goodman and Simon Lucas.
\newblock Does it matter how well {I} know what you're thinking? opponent
  modelling in an {RTS} game.
\newblock In {\em {IEEE} Congress on Evolutionary Computation, {CEC} 2020},
  pages 1--8. {IEEE}, 2020.

\bibitem[\protect\citeauthoryear{He \bgroup \em et al.\egroup
  }{2016}]{he2016opponent}
He~He, Jordan Boyd-Graber, Kevin Kwok, and Hal Daum{\'e}~III.
\newblock Opponent modeling in deep reinforcement learning.
\newblock In {\em International Conference on Machine Learning}, pages
  1804--1813, 2016.

\bibitem[\protect\citeauthoryear{Hernandez \bgroup \em et al.\egroup
  }{2019}]{hernandez2019generalized}
Daniel Hernandez, Kevin Denamgana{\"\i}, Yuan Gao, et~al.
\newblock A generalized framework for self-play training.
\newblock In {\em 2019 IEEE Conference on Games (CoG)}, pages 1--8. IEEE, 2019.

\bibitem[\protect\citeauthoryear{Hernandez-Leal \bgroup \em et al.\egroup
  }{2017}]{hernandez2017survey}
Pablo Hernandez-Leal, Michael Kaisers, Tim Baarslag, and Enrique~Munoz de~Cote.
\newblock A survey of learning in multiagent environments: Dealing with
  non-stationarity. arxiv 2017.
\newblock {\em arXiv preprint arXiv:1707.09183}, 2017.

\bibitem[\protect\citeauthoryear{Hernandez{-}Leal \bgroup \em et al.\egroup
  }{2019}]{hernandez2019agent}
Pablo Hernandez{-}Leal, Bilal Kartal, and Matthew~E. Taylor.
\newblock Agent modeling as auxiliary task for deep reinforcement learning.
\newblock In Gillian Smith and Levi Lelis, editors, {\em Proceedings of the
  Fifteenth {AAAI} Conference on Artificial Intelligence and Interactive
  Digital Entertainment, {AIIDE} 2019}, pages 31--37. {AAAI} Press, 2019.

\bibitem[\protect\citeauthoryear{Hong \bgroup \em et al.\egroup
  }{2018}]{hong2018deep}
Zhang-Wei Hong, Shih-Yang Su, Tzu-Yun Shann, et~al.
\newblock A deep policy inference q-network for multi-agent systems.
\newblock In {\em Proceedings of the 17th International Conference on
  Autonomous Agents and MultiAgent Systems}, pages 1388--1396. International
  Foundation for Autonomous Agents and Multiagent Systems, 2018.

\bibitem[\protect\citeauthoryear{Lanctot \bgroup \em et al.\egroup
  }{2017}]{lanctot2017unified}
Marc Lanctot, Vinicius Zambaldi, Audrunas Gruslys, et~al.
\newblock A unified game-theoretic approach to multiagent reinforcement
  learning.
\newblock {\em Advances in Neural Information Processing Systems},
  30:4190--4203, 2017.

\bibitem[\protect\citeauthoryear{Liu \bgroup \em et al.\egroup
  }{2021}]{liu2021towards}
Xiangyu Liu, Hangtian Jia, Ying Wen, et~al.
\newblock Towards unifying behavioral and response diversity for open-ended
  learning in zero-sum games.
\newblock {\em Advances in Neural Information Processing Systems}, 34, 2021.

\bibitem[\protect\citeauthoryear{Mnih \bgroup \em et al.\egroup
  }{2013}]{mnih2013playing}
Volodymyr Mnih, Koray Kavukcuoglu, David Silver, et~al.
\newblock {Playing Atari with Deep Reinforcement Learning}.
\newblock {\em CoRR}, abs/1312.5602, 2013.

\bibitem[\protect\citeauthoryear{Nashed and
  Zilberstein}{2022}]{nashed2022survey}
Samer Nashed and Shlomo Zilberstein.
\newblock A survey of opponent modeling in adversarial domains.
\newblock {\em Journal of Artificial Intelligence Research}, 73:277--327, 2022.

\bibitem[\protect\citeauthoryear{Oliehoek and Amato}{2014}]{oliehoek2014best}
Frans~A Oliehoek and Christopher Amato.
\newblock Best response bayesian reinforcement learning for multiagent systems
  with state uncertainty.
\newblock In {\em Proceedings of the Ninth AAMAS Workshop on Multi-Agent
  Sequential Decision Making in Uncertain Domains (MSDM)}, 2014.

\bibitem[\protect\citeauthoryear{Pascanu \bgroup \em et al.\egroup
  }{2013}]{pascanu2013difficulty}
Razvan Pascanu, Tomas Mikolov, and Yoshua Bengio.
\newblock On the difficulty of training recurrent neural networks.
\newblock In {\em International conference on machine learning}, pages
  1310--1318. PMLR, 2013.

\bibitem[\protect\citeauthoryear{Schaul \bgroup \em et al.\egroup
  }{2015}]{Schaul2015}
Tom Schaul, John Quan, Ioannis Antonoglou, and David Silver.
\newblock {Prioritized Experience Replay}.
\newblock {\em CoRR}, abs/1511.05952, 2015.

\bibitem[\protect\citeauthoryear{Schulman \bgroup \em et al.\egroup
  }{2017}]{Schulman2017}
John Schulman, Filip Wolski, Prafulla Dhariwal, et~al.
\newblock Proximal policy optimization algorithms.
\newblock {\em CoRR}, abs/1707.06347, 2017.

\bibitem[\protect\citeauthoryear{Silver \bgroup \em et al.\egroup
  }{2016}]{silver2016mastering}
David Silver, Aja Huang, Chris~J Maddison, et~al.
\newblock {Mastering the game of Go with deep neural networks and tree search}.
\newblock {\em Nature}, 529(7587):484--489, 2016.

\bibitem[\protect\citeauthoryear{Silver \bgroup \em et al.\egroup
  }{2018}]{Silver2017b}
David Silver, Thomas Hubert, Julian Schrittwieser, et~al.
\newblock {Mastering Chess and Shogi by Self-Play with a General Reinforcement
  Learning Algorithm}.
\newblock {\em Science}, 362:1140--1144, 2018.

\bibitem[\protect\citeauthoryear{Soemers \bgroup \em et al.\egroup
  }{2019}]{soemers2019learning}
Dennis~JNJ Soemers, Eric Piette, Matthew Stephenson, and Cameron Browne.
\newblock Learning policies from self-play with policy gradients and mcts value
  estimates.
\newblock In {\em 2019 IEEE Conference on Games (CoG)}, pages 1--8. IEEE, 2019.

\bibitem[\protect\citeauthoryear{Soemers \bgroup \em et al.\egroup
  }{2020}]{soemers2020manipulating}
Dennis~JNJ Soemers, {\'E}ric Piette, Matthew Stephenson, and Cameron Browne.
\newblock Manipulating the distributions of experience used for self-play
  learning in expert iteration.
\newblock {\em arXiv preprint arXiv:2006.00283}, 2020.

\bibitem[\protect\citeauthoryear{Timbers \bgroup \em et al.\egroup
  }{2022}]{timbers2022approximate}
Finbarr Timbers, Nolan Bard, Edward Lockhart, et~al.
\newblock Approximate exploitability: Learning a best response.
\newblock In {\em Proceedings of the International Joint Conference on
  Artificial Intelligence (IJCAI)}, pages 3487--3493, 2022.

\bibitem[\protect\citeauthoryear{Vinyals \bgroup \em et al.\egroup
  }{2019}]{vinyals2019grandmaster}
Oriol Vinyals, Igor Babuschkin, Wojciech~M Czarnecki, et~al.
\newblock Grandmaster level in starcraft ii using multi-agent reinforcement
  learning.
\newblock {\em Nature}, 575(7782):350--354, 2019.

\bibitem[\protect\citeauthoryear{Willemsen \bgroup \em et al.\egroup
  }{2021}]{willemsen2021value}
Daniel Willemsen, Hendrik Baier, and Michael Kaisers.
\newblock Value targets in off-policy alphazero: a new greedy backup.
\newblock {\em Neural Computing and Applications}, pages 1--14, 2021.

\bibitem[\protect\citeauthoryear{Wu}{2019}]{wu2019accelerating}
David~J Wu.
\newblock Accelerating self-play learning in go.
\newblock {\em arXiv preprint arXiv:1902.10565}, 2019.

\end{thebibliography}

\clearpage
\appendix

\section{Appendix}\label{section:appendix}

\subsection{Training \& benchmarking opponents}~\label{chapter2:section:training_test_agents}

We used $\delta=0$-Uniform self-play~\cite{hernandez2019generalized}, both for training our agents with BRExIt and its various ablations, and for training our test opponents with PPO. This self-play scheme makes a copy of the learning agent at the end of each episode, and adds it to its population -- the set of available policies to be used as training opponents. At the beginning of each training episode, a policy is randomly sampled from the population as opponent for the learning agent. This self-play scheme is effective for relatively simple games like Connect4, because it ensures that the learning agent continues to face all policies discovered during training, preventing catastrophic forgetting and cyclic policy evolution, which leads to transitively better strategies.

Initially, we generated three internally monotonically stronger test opponents $\bm{\pi^{test}} = [\pi_0, \pi_1, \pi_2]$, by freezing a copy of a PPO~\cite{Schulman2017} agent trained under $\delta=0$-Uniform self-play after 200k, 400k, and 600k episodes respectively. Table~\ref{table:PPOHyper} shows the hyperparameters used to train these
agents. No formal hyperparameter sweep was conducted, and the final values were
chosen after a few manual trials. We experimented with frame stacking to add a
temporal dimension to the state space, but it was ultimately discarded as
it lead to weaker agents for our compute budget.

\begin{table*}[tb]
    \centering
    \caption{PPO hyperparameters used to generate test opponents. Empty entries in rightmost column indicate that a single value was used.}
    \begin{tabular}{l c c}
        \toprule
        Hyperparameter name & Value & Values explored\\
        \midrule
        Horizon (T) & $2048$ & 512, 1024, 2048\\
        Adam stepsize & $10^{-5}$ & -\\
        Num. epochs & $10$ & 5, 10\\
        Minibatch size & $256$ & 128, 256, 512 \\
        Discount ($\gamma$) & $0.99$ & - \\
        GAE parameter ($\lambda$) & $0.95$ & - \\
        Entropy coefficient & $0.01$ & -\\
        Clipping parameter ($\epsilon$) & $0.2$ & - \\
        Gradient norm clipping & $1$ & 1, 5 \\
        \midrule
        Channels & [3, 10, 20, 20, 20, 1] & - \\
        Kernel sizes & [3, 3, 3, 3, 3] & - \\
        Paddings & [1, 1, 1, 1, 1] & - \\
        Strides & [1, 1, 1, 1, 1] & - \\
        Residual connections & [0,1], [1,2], [2,3], [3,4] & -\\
        \bottomrule
    \end{tabular}
    \label{table:PPOHyper}
\end{table*}

\begin{figure}[b!]
\centering
    \includegraphics[width=0.7\linewidth]{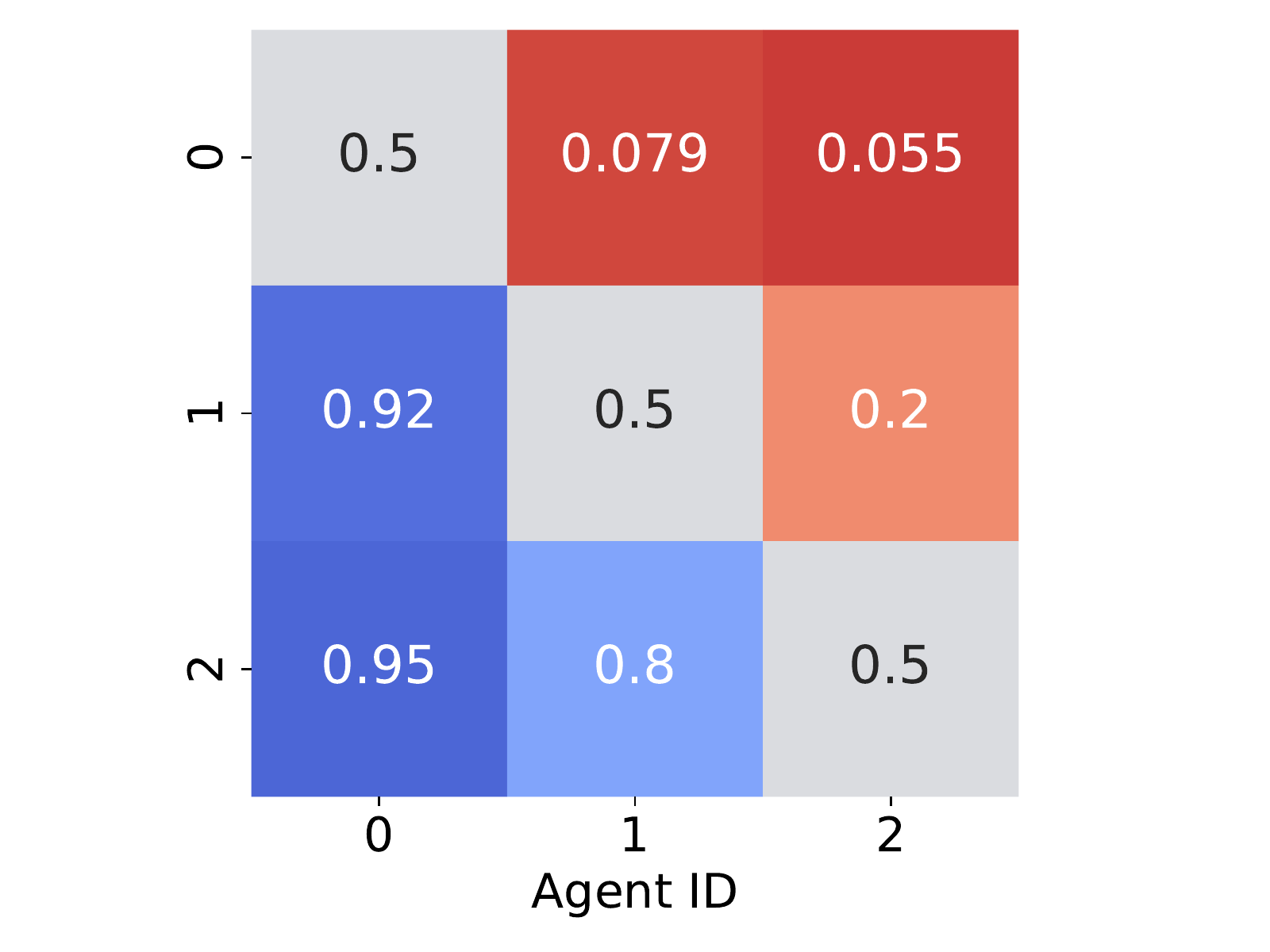}
    \caption{Winrates of $\pi_{\text{row}}$ against $\pi_{\text{column}}$, where each entry was computed by playing $1000$ head-to-head matches in the game of Connect4. The color of each agent (first player or second player) was chosen randomly at the beginning of each match to enforce symmetry. The agent IDs 0, 1 and 2 correspond to the 200K, 400K and 600K agents respectively.}
    \label{figure:winrate_matrix_test_agents}
\end{figure}

To evaluate the relative strength of the test opponents, we computed their corresponding winrate evaluation matrix to study pair-wise agent performances, as shown in Figure~\ref{figure:winrate_matrix_test_agents}. We see that later agents can consistently beat earlier versions. This shows a transitive improvement \textit{within} the population of agents that emerges from our single training run. However, the nature of PPO agents is reactive, as there is no planning involved during their training. This means that agents can be internally better because they learn how to exploit specific weaknesses of previous agents, which might not be present in unrelated agents, for example agents generated via planning based methods such as BRExIt. Because BRExIt uses planning to choose its actions during training, we ultimately want to give a ranking to our test agents against MCTS based methods. In order to label our test agents with the labels of \emph{weak} and \emph{strong} used in the main text, we use \textbf{MCTS equivalent strength}, which we define as the computational budget (number of MCTS iterations) required for an MCTS agent using random rollouts to reach a given winrate against a given agent policy $\pi$.

\begin{figure}[h]
\centering
    \includegraphics[width=\linewidth, keepaspectratio]{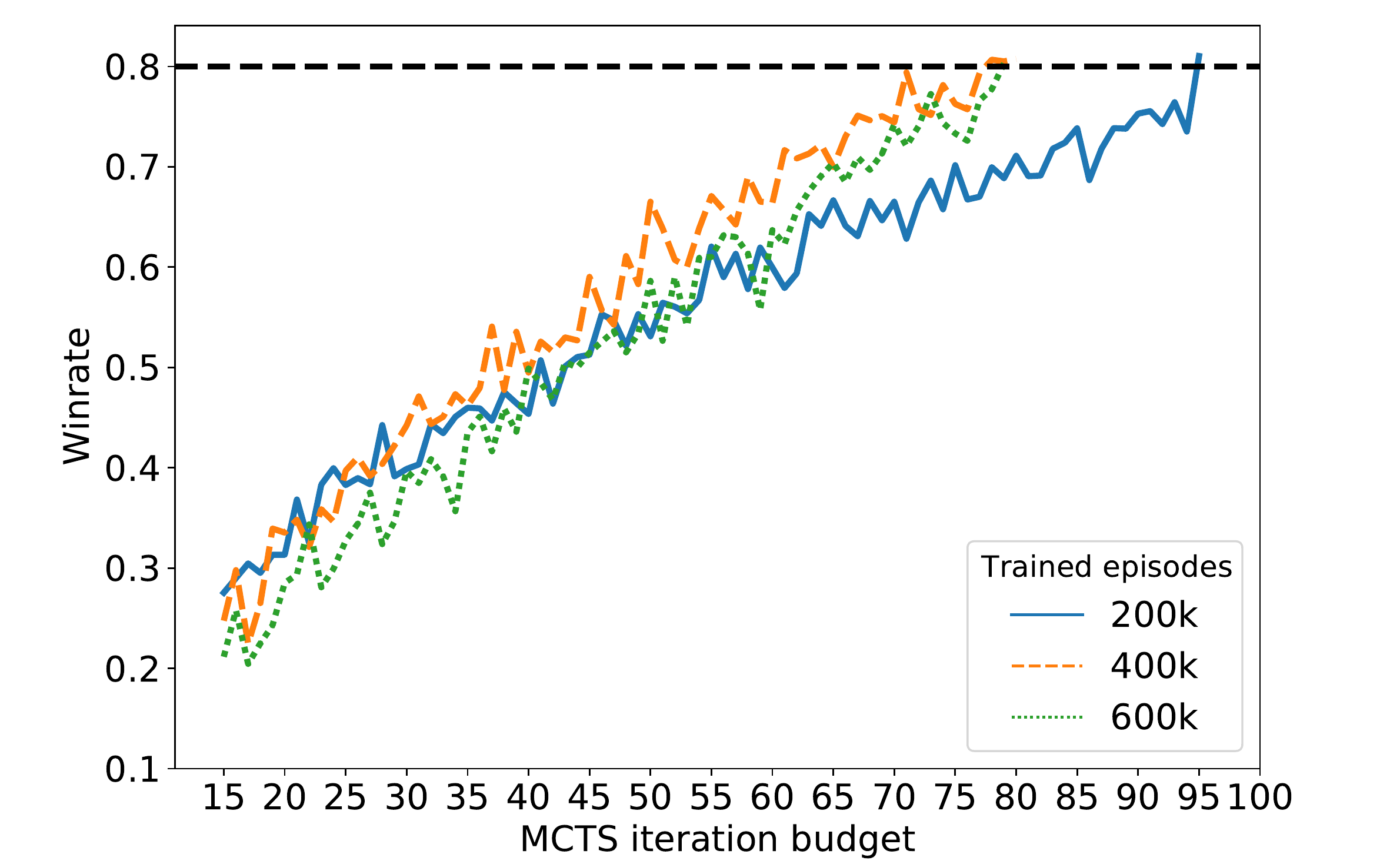}
    \caption{
Evolution of winrate by MCTS against all candidate test agents (excluding $\pi_{mixed}$). As the MCTS computational budget increases on the horizontal axis, so does its playing strength. The black horizontal dashed line at the $80\%$ winrate mark is shown to compare the required budget to reach such winrate against all test opponents.}
    \label{figure:mcts_equivalent_strength}
\end{figure}

Figure~\ref{figure:mcts_equivalent_strength} shows a sweep of MCTS equivalent strength up to $80\%$ winrate. Surprisingly, we see that the test agent with only 200K training episodes, even though it is the weakest when matched directly against other agents from the test population as seen in Figure~\ref{figure:winrate_matrix_test_agents}, is the strongest against MCTS with an MCTS equivalent strength of 95. We hypothesize that in the relatively low computational budgets that we have used for the agents trained in Section~\ref{section:experiments} of the main text, the heavily stochastic behaviour of the 200K agent can thwart shallow plans made by low budget MCTS agents. The MCTS equivalent strength for both other agents is 79. Given the superiority of the agent trained for 600K episodes in Figure~\ref{figure:winrate_matrix_test_agents} when compared to the agent trained for 400K, and their equivalent strengths against MCTS, we ultimately decided to discard the 400K agent as a test opponents for our main experiments. It is from this strength analysis against MCTS that we label the 600K and 200K agents as \textit{weak} and \emph{strong} respectively in the main paper.

\subsection{Neural network architectures}

In order to level the playing field for all algorithms, we adjusted
the actor-critic architecture, used by ExIt, and the opponent modelling
enhanced actor-critic architecture, used by BRExIt and its ablations, to feature a similar
number of parameters (approximately 27,000 trainable parameters).

The game of Connect4 is played on a $6 \times 7$ grid, and each grid position
can either be empty, occupied by one of player 1's chips, or by one of player
2's chips. The input to PPO's and BRExIt's neural network models was a $3 \times 6
\times 7$ tensor (a 3 channeled $6 \times 7$ grid): The first channel features
$0$s on all non-empty positions and $1$s on empty positions; the second channel
features $1$s on positions taken by the current player and $0$s everywhere else; the third channel features $1$s on positions taken by the opponent and $0$s everywhere else.

We used a similar network architecture to~\cite{hernandez2019generalized}. The input tensor was put through 5 convolutional layers, with residual connections skipping every other layer. Extra details are shown in Table~\ref{table:PPOHyper}. The output embedding found at the end of the last convolutional layer was fed into two parallel layers. One was a fully connected layer with a softmax activation function which represented the agent's policy (the actor). The other was a single neuron and no activation function representing a state-value function (the critic).

\begin{table*}[h]
    \centering
    \caption{BRExIt, BRExIt-OMS, ExIt and ExIt-OMFS hyperparameters. Hyperparameter tuning was done manually and in a limited way due to limited computational resources. Empty entries in the rightmost column indicate that a single value was used. The neural network architecture shared across all agents was adapted from the original DPIQN~\protect\cite{hong2018deep}.}
    \begin{tabular}{l c c c}
        \toprule
        Hyperparameter name & Value & Values explored\\
        \midrule
        MCTS budget & 50 & 30, 50, 80 \\
        MCTS rollout budget & 0 & $\infty$, 0 \\
        MCTS exploration factor & 2. & $1, 2, 5$ \\
        Dirichlet noise & False & True, False\\
        Batch size & 512 & 128, 256, 512 \\
        Espisodes per generation (train every $n$ episodes) & 800 & 300, 500, 800, 1000\\
        Epochs per iteration & 5 & 1, 3, 5, 10\\
        Reduce temperature after $n$ moves & 10 & $\infty$, 10 \\
        \midrule
        Channels & [12, 15, 20, 20, 20, 1] & -\\
        Kernel sizes & [3, 3, 3, 3, 3] & -\\
        Paddings & [1, 1, 1, 1, 1] & -\\
        Strides & [1, 1, 1, 1, 1] & -\\
        Residual connections & [[0,1], [1,2], [2,3], [3,4]] & -\\
        \midrule
        Learning rate & 1.5e-3 & 1e-3, 1.5e-3\\
        Post feature extractor hidden units & [128, 128, 128] & -\\
        Post feature extractor policy inference hidden units & [128, 64] & -\\
        Post feature extractor actor critic hidden units & [128, 64] & -\\
        Activation function & \textit{ReLu} & -\\
        Critic activation function & \textit{tanh} & -\\
        Gradient norm clipping & 1 & None, 1, 5 \\
        \bottomrule
    \end{tabular}
    \label{chapter2:table:brexit_hyperparameters}
\end{table*}

\subsection{Kolmogorov-Smirnov test}\label{section:kolmogorov_smirnov_tests}

Table~\ref{table:p_values_kolmogorov_smirnov_tests} in Section~\ref{section:experiments} shows the $p$-values corresponding to the different two-sided Kolmogorov-Smirnov tests. This test measures if there is a significant statistical difference between 2 distributions $f(x)$ and $g(x)$. In our case, $f, g$ correspond to an algorithmic ablation (i.e BRExIt, BRExIt-OMS and ExIt-OMFS), with $g$ representing the same ablation as $f$ but using one hot action encodings for OM targets instead of full action distributions. $x$ denotes a test agent policy ($\pi_{weak}, \pi_{strong}$ or $\pi_{mixed}$). Hence, $f(x), g(x)$ denote the stochastic functions that determine the final winrate obtained by $f$ or $g$ respectively at the end of a training run against policy $x$. Each of the 6 training runs used for each algorithmc ablation (originally 10, but we discard the top and bottom 2 performing runs as per~\cite{agarwal2021deep}) gives us one data point. Thus, our comparisons are between sets of 6 datapoints $f(x) = [f_1(x), f_2(x), f_3(x), f_4(x), f_5(x), f_6(x)]$ and similarly $g(x) = [g_1(x), g_2(x), g_3(x), g_4(x), g_5(x), g_6(x)]$.

\subsection{Hyperparameters and general performance improvements}\label{section:brexit_hyperparameters_and_performance_improvements}

Deep reinforcement learning algorithms require a large amount of simulated
experiences to train policies, and more so for deep multi-agent reinforcement learning. This is further
exacerbated when planning based algorithms are used as part of RL algorithms,
as is the case with MCTS within BRExIt. A single step in the main game requires
many simulated games inside of MCTS. On top of this, RL methods do not tend to be robust to hyperparameters, requiring a lot of manual tuning and
stabilizing implementation details. Because of this, it is common place to add
ad-hoc or environment specific methods to reduce the amount of compute required
to train policies and to increase their robustness. Here we present some
performance improvements shared across all algorithmic ablations, which are
orthogonal to the novel contributions presented in this paper but were
nonetheless required to train our agents within our computational budget. We
argue that these improvement methods benefit all ablations equally, and thus do
not affect our comparisons. We present them here for completeness and to aid
reimplementations efforts. The set of hyperparameters used for all algorithmic
ablations during the final training run is present in
Table~\ref{chapter2:table:brexit_hyperparameters}.

\subsubsection{Reducing exploration: removing Dirichlet noise on MCTS root priors}

The algorithm AlphaGo~\cite{Silver2017b} introduced the notion of adding noise
to the prior probabilities in the root node. By adding noise, we modify the
priors over actions used in the selection phase, given by the apprentice's
state-value function. This is used as an exploration mechanism, by
redistributing some probability weight given by the priors over all valid
actions. The Dirichlet distribution is used to sample that noise: $P(s, a) =
0.25 * Dirichlet(\alpha) + 0.75 * P(s, a)$, where $\alpha = \sqrt{2}$. During an MCTS search with many hundreds of iterations (as was the case with AlphaZero), the explorative effects of this Dirichlet noise on the root node's priors would eventually be washed out by state evaluations propagated upwards from further down the tree. In our case however, due to our relatively low MCTS budget of 50, we cannot afford to use Dirichlet noise, as this exploratory noise would randomize the output policy of the search $\pi_{MCTS}$ too much.

\subsubsection{Sample efficiency: Data augmentation via exploiting state space symmetries}

The board of Connect4, a $6 \times 7$ matrix, is symmetrical over the vertical
axis. We can exploit this to reduce the complexity of the environment. Let's
define a function over states $\sigma(\mathcal{S}) \rightarrow
\mathcal{S}$ which swaps columns with indexes $[1, 2, 3]$ with those with
index $[7, 6, 5]$; and also over policies $\sigma(\Pi) \rightarrow \Pi$ by
also swapping the probability weight over those column indices. Strategically
speaking, the swapped situations are identical. AlphaZero was shown to be able to naturally learn this symmetry over enough training episodes~\cite{Silver2017b}. We add this symmetry a priori into our algorithm to reduce computational costs. This is done by augmenting every data point at the time it is appended to the replay buffer by
also adding a copy with the corresponding symmetric transformation, on both the
state and the policy target (and the opponent policy target in case of
BRExIt). The value target remains the same. Formally, for every data point
we obtain a tuple:

\begin{align*}
&\{s_t, \pi_{MCTS}(\cdot | s_t), G^i_t\} \rightarrow \\ &(\{s_t, \pi_{MCTS}(\cdot |
s_t), G^i_t\}, \{\sigma(s_t), \sigma(\pi)_{MCTS}(\cdot | s_t), G^i_t\})
\end{align*}

\subsubsection{Variance reduction on value targets: Averaging MCTS estimated returns with episodic returns}

In the original ExIt specification, as described in
Equation~\ref{equation:exit_dataset}, at every timestep of each training
episode, the following information is stored in a replay buffer: $\{s_t,
\pi_{MCTS}(\cdot | s_t), G^i_t\}$, where $G^i_t$ denotes player $i$'s total
reward observed at the end of the episode. What this means for our state value
function $V(s_t)$ is that we are regressing all observed states in an episode
$[s_0, s_1, \ldots]$ to the same value $G^i_t$. This target features a large variance, as we are mapping potentially dozens of states to the same target
value. We can reduce the variance, with the downside of introducing some bias, by
re-using state-dependent computation from each state-dependent MCTS call,
meaning that we don't use the same value target for each state anymore. For every
data point we substitute $G^i_t$ by another variable $z^i_t$, for which we tried
three different alternatives:

\begin{equation}
    z^i_t = \frac{1}{2} * (\underbrace{\sum_{a \in \mathcal{A}} \pi_{MCTS}(a | s) * Q(s, a)}_\textrm{Definition of $V^{\pi_{MCTS}}(s)$} + G^i_t)
\end{equation}

First, we averaged the original target with the root node value of MCTS. This includes the value of sub-optimal actions that were sampled during exploration. This
in turn induced pessimism in $V(s)$ estimations by our trained critics
$V_{\bm{\phi}}(s)$, leading them to consistently estimate lower winrates than
observed in practice.

\begin{equation}
    z^i_t = \frac{1}{2} * (\underbrace{\text{max}_{a \in \mathcal{A}} * Q(s, a)}_\textrm{Greedy w.r.t $Q(s, \cdot)$} + G^i_t)
\end{equation}

Second, we tried using the value of the root node's highest valued child node. This
also introduced a bias, although this time in a positive direction, because the
policy that will actually be followed corresponds to MCTS's normalized child visitations (Equation~\ref{equation:pi_mcts}) -- which will put some probability weight on
lower valued child nodes as well.

\begin{equation}
    z^i_t = \frac{1}{2} * (\underbrace{Q(s, \argmax_{a \in \mathcal{A}} N(s, a))}_\textrm{MCTS action selection} + G^i_t)
\end{equation}

Ultimately, we decided to average the episode returns with the $Q(s,a)$ value
associated with the action actually selected and returned by MCTS. In our experiments, this was the most visited action at the root. Theoretically
this still yields biased targets, as our actors $\pi_{\bm{\theta}}$ imitate an
MCTS derived policy $\pi_{MCTS}$ while our critic $V_{\bm{\phi}}$ will regress
against state-values that do not exactly match the actions taken by
$\pi_{MCTS}$. However, not only did this seem to work well in practice, but we saw
little to no bias in our critic's estimations compared to the real episode
returns using this bootstrapped estimate.

\subsubsection{Extra exploitation: Temperature parameter}

It is beneficial to have a high degree of exploration on the initial states of
an environment, specially during early stages of training. The policy targets derived from normalized child visits at the root node $\pi_{MCTS}$ feature a high degree of exploration, especially with our low computational budget, as MCTS would discard low quality moves given larger computational budgets. This is convenient early during training for the aforementioned exploration on initial states, but can slow down the learning of
exploitative policies against fixed agents. Thus, ideally we would like to focus initial moves on exploration, and switch to exploitation later into an episode.

We achieve this with temperature parameter $\tau$, which we
use to exponentiate the visits to root node children before they are
normalized:

\begin{equation}
    \pi_{MCTS}(a | s) = \frac{N(s,a)^{1/\tau}}{\sum_{a'} N(s, a')^{1/\tau}}
\end{equation}

We set $\tau = 1$ for the first 10 moves, after which we reduce it to $\tau =
0.01$, which effectively moves all probability weight to the most visited
action. Therefore, the first 10 moves will feature increased exploration, and all other moves will focus on the most-visited action. We note that there exists research on this area, with a focus on removing exploration elements from MCTS policy targets with the hope of aiding interpretability~\cite{soemers2019learning}.

\subsubsection{Gradient norm clipping}

Many RL algorithms which involve computing gradients with respect to a loss
function suffer from high variance in the estimation of these gradients. If the
parameters of a model are updated by applying gradients of a large magnitude,
these might move the model's parameters too far, even into poor performing sections of the
parameter space. A proposed solution is to introduce gradient
clipping~\cite{pascanu2013difficulty}. Its most naive implementation involves
clipping the value of \textit{each individual} gradient to a hyperparameter
threshold $c$. A more nuanced alternative is gradient norm clipping. It
involves concatenating all gradients of all parameters into a single vector
$\bm{g}$. If the vector's $n$th norm is greater than a threshold value $c$,
then the vector is normalized according to:
\begin{equation}
    \bm{g} = 
    \begin{cases}
        \bm{g} \quad & \text{if} ||\bm{g}||_n \leq c \\
        c\frac{\bm{g}}{||\bm{g}||_n} & \text{otherwise} \\
    \end{cases}
\end{equation}
For all experiments we used the $L_2$ norm. We tried thresholds of $5$, $3$ and
$1$. Even though we found that higher thresholds yielded a sharper increase in
performance early on, a threshold of $c=1$ ended up achieving better stability
and final performance than the other values.

%
%

\subsubsection{Parallel data generation: Parallelized MCTS with centralized apprentice}

We simulate multiple games of Connect4 in parallel, With every MCTS procedure
running on an individual CPU core. A single process hosts the apprentice's
neural network, acting as a server, whose only function is to receive
states from all MCTS processes. These are evaluated for (1) node priors
$P(s, a)$ and (2) state evaluations $V(s)$. It busy polls all connections to
MCTS processes for incoming states, batching all requests into a single
neural network forward pass. This allows us to scale horizontally,
theoretically linearly with the number of available CPUs. Because of the size
of the network, Connect4's small state size and the average batch size,
we saw no performance improvement hosting the apprentice in CPU vs GPU.
Surprisingly, our main bottleneck came from Python's communication of PyTorch
tensors among processes, stalling MCTS speed while requesting node evaluations $V(s)$ and node prior generations $P(s, a)$.

\subsection{Description of computational infrastructure}

The experiments presented in this publication were carried out in Snellius, the Dutch National supercomputer, a SLURM based computing cluster. Each of the 280 runs from Section~\ref{section:experiments} were individual SLURM jobs that run for 48h. They each had access to 1280GB of RAM and 64 CPUs with no GPUs.

\end{document}